\documentclass{article}

\usepackage[nonatbib,preprint]{neurips_2021}

\usepackage[utf8]{inputenc} 
\usepackage[T1]{fontenc}    
\usepackage{hyperref}       
\usepackage{url}            
\usepackage{booktabs}       
\usepackage{amsfonts}       
\usepackage{nicefrac}       
\usepackage{microtype}      
\usepackage{xcolor}         

\usepackage[utf8]{inputenc}

\usepackage{color}

\usepackage[
maxcitenames=3, 
maxbibnames=10,  
sorting=none,
sortcites,
defernumbers,
backend=biber
]{biblatex}
\usepackage{hyperref}
\hypersetup{
    final=true, 
    plainpages=false,
    pdfstartview=FitV,
    pdftoolbar=true,
    pdfmenubar=true,
    pdfencoding=auto,
    psdextra,
    bookmarksopen=true,
    bookmarksnumbered=true,
    breaklinks=true,
    linktocpage=true,
 	pdfinfo={
 	    Title={Neural Architecture Search via Inverse Scale Space Iterations},
	    Keywords={Neural Architecture Search, Sparse Neural Networks, Inverse Scale Space}
	    author={Leon Bungert, Tim Roith, Daniel Tenbrinck, Martin Burger}}
}
\usepackage{amsmath,amssymb,amsthm}

\usepackage[ruled,vlined]{algorithm2e}

\usepackage[capitalise,noabbrev,nameinlink]{cleveref}
\crefname{assumption}{Assumption}{Assumptions}

\usepackage{caption}
\usepackage{subcaption}

\usepackage{todonotes}
\usepackage{tcolorbox}

\usepackage{booktabs}
\usepackage{multirow}
\usepackage{hhline}

\usepackage{tabularx}
\newcolumntype{R}{>{\raggedright\arraybackslash}X}
\newcolumntype{C}{>{\centering\arraybackslash}X}
\newcolumntype{L}{>{\raggedleft\arraybackslash}X}

\definecolor{internationalkleinblue}{rgb}{0.0, 0.18, 0.65}

\newcommand{\R}{\mathbb{R}}

\DeclareMathOperator*{\argmin}{arg\,min}

\newcommand{\prox}[1]{\,\mathrm{prox}_{#1}}

\newcommand{\norm}[1]{\Vert#1\Vert}
\newcommand{\st}{\,:\,}

\DeclareMathOperator*{\sign}{sign}

\newcommand{\param}{\theta}

\newcommand{\trSet}{\mathcal{T}}

\newcommand{\empBatchLoss}{L}

\newcommand{\func}{J}

\newcommand{\AdaBreg}{\mbox{AdaBreg}}

\theoremstyle{remark}

\numberwithin{equation}{section}

\title{Neural Architecture Search \\via Bregman Iterations}

\author{%
  Leon Bungert \qquad
  Tim Roith \qquad 
  Daniel Tenbrinck \qquad
  Martin Burger\\
  Department Mathematics,
  Friedrich-Alexander University Erlangen-Nürnberg, Germany\\
  \texttt{\{leon.bungert, tim.roith, daniel.tenbrinck, martin.burger\}@fau.de} 
}

\makeatletter
\let\blx@rerun@biber\relax
\makeatother

\begin{document}

\maketitle

\begin{abstract}
  We propose a novel strategy for Neural Architecture Search (NAS) based on Bregman iterations. 
  Starting from a sparse neural network our gradient-based one-shot algorithm gradually adds relevant parameters in an inverse scale space manner.  
  This allows the network to choose the best architecture in the search space which makes it well-designed for a given task, e.g., by adding neurons or skip connections.
  We demonstrate that using our approach one can unveil, for instance, residual autoencoders for denoising, deblurring, and classification tasks. 
  Code is available at \url{https://github.com/TimRoith/BregmanLearning}.
\end{abstract}

\section{Introduction}

The success story of Deep Neural Networks (DNNs) in fields like computer vision, image and language processing, etc., can be attributed to different factors:
First, DNNs are highly overparametrized nonlinear mappings which therefore possess large expressiveness. 
Second, increasingly sophisticated architecture choices, most prominently convolutional layers and residual connections, resembling biological processes have proved to be adequate models for catching up to or even surpassing human performance on tasks like image classification \cite{he2016deep}, segmentation \cite{ronneberger2015u}, or gaming \cite{silver2017mastering}. 

Their overparametrization is both boon and bane of DNNs since it may cause instabilities and overfitting, and makes the networks difficult to train \cite{Goodfellow16}. 
Mitigating these drawbacks is an active field of research and popular approaches aim to train sparse neural networks \cite{scardapane2017group,dettmers2019sparse,hoefler2021sparsity,frankle2018lottery,zhu2017prune}, which are both less prone to overfitting, more memory-efficient, and facilitate efficient sparse training.
Also the usage of more advanced optimization methods like stochastic gradient descent with momentum, Adam or related methods \cite{qian1999momentum,kingma2014adam,orvieto2020role,ruder2016overview} empirically tends to produce more stable networks.

Seemingly unrelated to sparsity and optimization is the question which network architectures work best for a specific application.
However, as we discuss below, these topics are strongly related. 
Handcrafted architectures like the VGG networks \cite{simonyan2014very}, residual neural networks (ResNets) \cite{he2016deep}, or U-Nets \cite{ronneberger2015u} perform extremely well for various applications.
Furthermore, more recently designed architectures like, e.g., dense convolutional networks \cite{huang18} or Momentum ResNets \cite{sander21} suggest that network design has not yet reached its limit.
However, designing successful architectures manually is very time-consuming and at the same time it is not entirely clear how much influence the specific architecture has on the great performance of these networks. 

To obtain more insights on this question, the vibrant field of Neural Architecture Search (NAS) aims to let networks ``learn their own architecture'', cf. \cite{elsken2019neural} for a detailed review with many important references.
The different approaches for NAS can be coarsely subdivided based on which \emph{search space} and which \emph{search strategy} are used.
Common search spaces comprise simple chain-structured networks, more complicated networks with skip connections and branching, convolutional neural networks, etc.
Popular search strategies include random search, evolutionary algorithms, Bayesian or gradient-based optimization, and reinforcement learning.

\section{The Method}\label{sec:method}

\begin{figure}
    \def\PicWidth{0.24\textwidth}
    \centering
    {\bf Denoising} \hspace{5.5cm} {\bf Deblurring}\\
    {\includegraphics[width=\PicWidth,trim=0.3cm 0.3cm 0.2cm 0.2cm,clip]{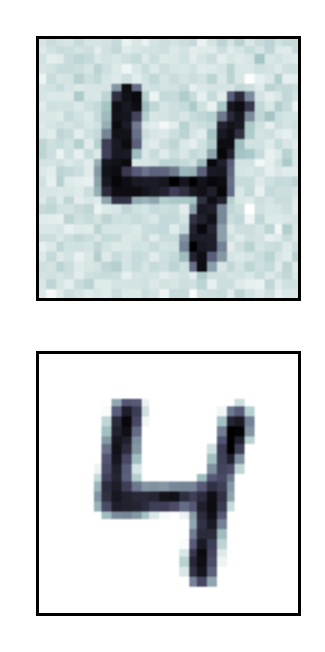}}%
    \hfill%
    \includegraphics[width=\PicWidth,trim=0.3cm 0.3cm 0.2cm 0.2cm,clip]{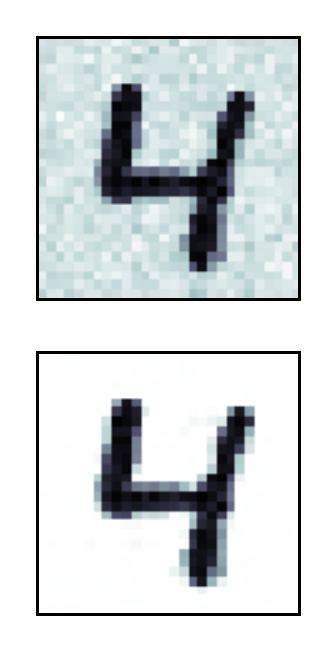}%
    \hfill%
    \includegraphics[width=\PicWidth,trim=0.3cm 0.3cm 0.2cm 0.2cm,clip]{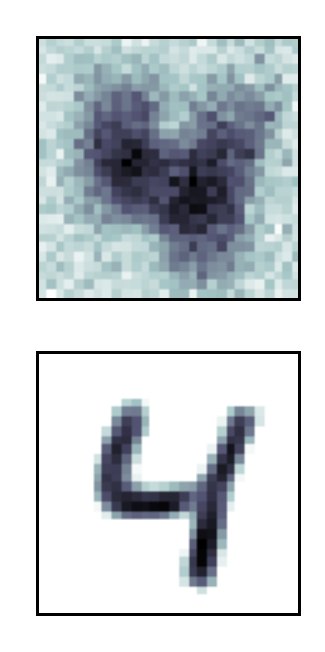}%
    \hfill%
    \includegraphics[width=\PicWidth,trim=0.3cm 0.3cm 0.2cm 0.2cm,clip]{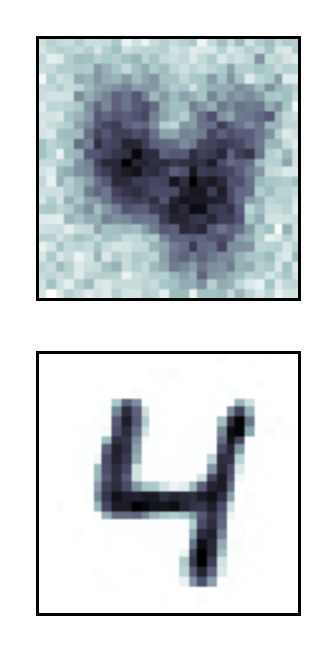}%
    \caption{Denoising and deblurring of MNIST test images with unveiled different architectures. \textbf{First and third image:} Autoencoder, \textbf{Second and fourth:} Residual autoencoder.}
    \label{fig:introimage}
\end{figure}

In this article we propose a novel gradient-based one-shot Neural Architecture Search method using Bregman iterations for training sparse neural networks \cite{bungert2021bregman}.
Bregman iterations are a generalized gradient descent algorithm and were originally proposed in \cite{osher2005iterative,burger2006nonlinear,cai2009linearized,burger2007inverse} for applications in compressed sensing and image processing. 
Their main characteristic is that they start with a sparse solution and successively add features to minimize a loss function, this way forming an \emph{inverse scale space}.
As we demonstrate in this article this special feature makes them a well-suited algorithm for NAS, additionally being supported with a rich mathematical theory \cite{burger2013adaptive,cai2009convergence,benning2018choose,benning2018modern,bungert2019nonlinear}.

\paragraph{Search Strategy}
Our \emph{search strategy} for discovering suitable neural network architectures is gradient-based and utilizes the algorithm \emph{AdaBreg} \cite{bungert2021bregman}, an inverse scale space variant of Adam~\cite{kingma2014adam}.
\begin{algorithm}[htb]
\def\commentWidth{7cm}
\newcommand{\atcp}[1]{\tcp*[r]{\makebox[\commentWidth]{#1\hfill}}}
\DontPrintSemicolon
\SetKwInOut{Input}{input}\SetKwInOut{Output}{output}
\SetKwInOut{Default}{default}
\Default{$\eta=0.001$,\quad$\beta_1=0.9$,\quad$\beta_2=0.999$,\quad$\epsilon=10^{-8}$,\quad$\delta=1$}
$k=0$,\quad $m_1 \gets 0$,\quad $m_2 \gets 0$ \atcp{initialize}
\For{epoch $e = 1$ \KwTo $E$}{
\For{minibatch $B\subset \trSet$}{
\tcp{Standard Adam steps}
$k \gets k+1$\;
$g \gets \nabla\empBatchLoss(\param;B)$\;
$m_1 \gets \beta_1\,m_1 + (1-\beta_1)\,g$\;
$\hat m_1 \gets m_1/(1-\beta_1^k)$\;
$m_2 \gets \beta_2\,m_2 + (1-\beta_2)\,g^2$\;
$\hat m_2 \gets m_2/(1-\beta_2^k)$\;
\tcp{Bregman steps}
$v\gets v - \eta\,\hat m_1/(\sqrt{\hat m_2} + \epsilon)$ \atcp{Moment step for architecture}
$\param \gets \prox{\delta\func}\left(\delta v\right)$ \atcp{Shrinkage step for active parameters}
}
}
\caption{\emph{\AdaBreg{}} \cite{bungert2021bregman}, a Bregman variant of Adam \cite{kingma2014adam} for Neural Architecture Search.}
\label{alg:adabreg}
\end{algorithm}

Here, $\eta>0$ is the learning rate and $\delta>0$, which we set equal to one, is a hyperparameter steering the magnitude of the network parameters $\param$. 
Furthermore, $\trSet$ denotes a training set and $L(\param;B)$ is a batch loss function of the network parameters $\param$.
The functional $\func$ is \emph{sparsity-enforcing}, and $\prox{}$ denotes the \emph{proximal operator} \cite{parikh2014proximal}, defined as
\begin{align}
    \prox{\func}(v) = \argmin_{w} \frac{1}{2}\norm{w-v}^2 + \func(w).
\end{align}
When choosing the functional $\func(\param)=0$, this is the identity operator and \cref{alg:adabreg} reduces to the popular Adam optimization method~\cite{kingma2014adam}.
The algorithm features two variables $v$ and $\param$ which are linked through the condition that $v$ is a subgradient of the elastic net \cite{zou2005regularization} functional
\begin{align}
\func_\delta(\param):=\func(\param)+\frac{1}{2\delta}\norm{\param}^2    
\end{align}
i.e., $v\in\partial\func_\delta(\param)$.
In the case that $\func(\param)=\norm{\param}_1$ equals the $\ell_1$-norm, the subgradients basically encode the \emph{support} of non-zero parameters since $v=\sign(\param)+\frac{1}{\delta}\param$.

\cref{alg:adabreg} first updates the variable $v$ by performing a gradient descent step on the loss function $L(\theta;B)$ and successively recovers only relevant parameters $\param$ by applying the proximal operator, which for $\func(\param)=\mu\norm{\param}$ reduces to \emph{soft-thresholding} \cite{parikh2014proximal} with parameter $\mu>0$:
\begin{align}
    \prox{\delta\mu\norm{\cdot}_1}(\delta v) = \delta \operatorname{shrink}(v;\mu) := \delta \sign(v)\max(|v|-\mu,0).
\end{align}
Since the threshold $\mu$ does not change over the iterations, \cref{alg:adabreg} exhibits an inverse scale space behavior by adding more and more parameters as soon as the corresponding entries of $v$ exceed it.

\paragraph{Search Space}
For the proposed Neural Architecture Search method the functional $\func$ typically involves a $\ell_{1,2}$ group-norm of different parameter groups of the network, first proposed in \cite{scardapane2017group}: 
\begin{align}\label{eq:group_sparsity}
    \func(\param) = \sum_{\mathbf{g}\in\mathcal{G}} \sqrt{n_\mathbf{g}}\norm{\mathbf{g}}_2.
\end{align}
Here, $\mathcal G$ denotes the collection of all parameter groups, such as weight matrices, convolutional kernels, skip connections, etc., and $n_\mathbf{g}$ denotes the number of elements of a group member $\mathbf{g}$, which ensures a balanced influence of all groups based on their number of parameters.
The norm $\norm{\cdot}_2$ denotes a suitable $\ell_2$-norm of the group member $\mathbf{g}$, which depending on the data structure can be the Euclidean norm of a vector, the Frobenius norm of a matrix, etc.

Note that we use the argument $\param$ in \eqref{eq:group_sparsity} as a collective variable containing all free parameters of the network.
The parameter groups $\mathcal{G}$, which are regularized through \eqref{eq:group_sparsity}, together with the parameter groups not entering the regularizer constitute the \emph{search space} of our NAS method.

As a prototypical example, let us study the following dense residual network structure with $L$ layers, where $x^{(0)}\in\R^d$ denotes the $d$-dimensional input and the output of the $l+1$-st layer is given by 
\begin{align}\label{eq:dense_resnet}
    x^{(l+1)} = \sigma(W^{(l)} x^{(l)} + b^{(l)}) + \sum_{i=0}^{l} \tau^{(l,i)} x^{(i)}, \quad l\in\{0,\dots,L-1\}.
\end{align}
Here, $W^{(l)}\in\R^{d\times d}$ denote weight matrices, $b^{(l)}\in\R^d$ are bias vectors, and $\tau^{(l,i)}\in\R$ are residual skip parameters. 
For suitable choices of the skip parameters, weights, and biases the architecture \eqref{eq:dense_resnet} covers special cases like standard feedforward network, ResNets or Neural ODE \cite{he2016deep,chen2018neural}, momentum ResNets \cite{sander21}, residual autoencoders \cite{dong2018learning}, etc.

Possible parameter groups of the model \eqref{eq:dense_resnet} are the row vectors of all weight matrices
\begin{align}
\mathcal{G}_\mathrm{neur}=\{W^{(l)}_{r,:}\st l=0,\dots,L-1,\,r=0,\dots,d-1\},  
\end{align}
the collection of all skip parameters
\begin{align}
\mathcal{G}_\mathrm{skip}=\{\tau^{(l,i)}\st l=0,\dots,L-1,\,i=0,\dots,l\},
\end{align}
or the layers of the network
\begin{align}
    \mathcal{G}_\mathrm{layer}=\{W^{(l)}\st l=0,\dots,L-1\}.
\end{align}
Obviously, these group can also be combined, leading to simultaneous regularization of different parameter groups.

Regularizing the first group enforces a small number of active neurons, which makes it a suitable model for discovering autoencoder-like architectures.
Similarly, the second group enforces few active residual connections and the third group few active layers.
Note that enforcing the latter is only meaningful when allowing for residual connections since otherwise the network does not transfer information.

As an example, consider the following two regularizers which enforce row sparsity with no skip connections, and joint sparsity of rows and skip connections, respectively:
\begin{align}
    \label{eq:regularizer_rows}
    \func(\param) &= \mu\sum_{l=0}^{L-1}\left(\sum_{r=0}^{d-1}\sqrt{d}~\norm{W^{(l)}_{r,:}}_2 + \sum_{i=0}^l \chi_{\{0\}}(\tau^{(l,i)})\right),\\
    \label{eq:regularizer_skip_rows}
    \func(\param) &= \mu\sum_{l=0}^{L-1}\left(\sum_{r=0}^{d-1}\sqrt{d}~\norm{W^{(l)}_{r,:}}_2 + \sum_{i=0}^l |\tau^{(l,i)}| \right),
\end{align}
where $\mu>0$ is a regularization parameter.
Note that by using the characteristic function $\chi_{\{0\}}(\tau)$---which is zero if $\tau=0$ and $\infty$ otherwise---the first regularizer forces all skip parameters to zero, whereas the second one just penalizes their $\ell_1$-norm.
For ResNet-type architectures one could additionally include a positivity constraint for the skip connection parameters $\tau^{(l,i)}$, whereas for Momentum ResNet-type architectures positivity does not have to be enforced.

\paragraph{Parameter Initialization}

We combine the proposed NAS method, which utilizes a suitable sparsity regularization as described before, with a sparse parameter initialization strategy, c.f., \cite{bungert2021bregman,liu2021,dettmers2019sparse}.
This allows our method to unveil its own architecture as it is not predetermined by the initialization. 
For this we choose an initial sparsity level $r\in[0,1]$ for each parameter group over which we optimize and initialize their group members in the following masked way:
\begin{align*}
    \mathbf{g} = \Tilde{\mathbf{g}}\cdot m,\quad m \sim \mathfrak{B}(r).
\end{align*}
Here, $m$ is a random variable drawn from a Bernoulli distribution $\mathfrak{B}(r)$ with parameter $r\in[0,1]$, and $\Tilde{\mathbf{g}}$ is a non-sparse random initialization.
For the latter, we either follow standard initialization techniques like \cite{bengio10,he2015delving} or utilize a modification proposed for sparse initialization in \cite{bungert2021bregman}, which rescales the variance of $\Tilde{\mathbf{g}}$ with ${1}/{r}$.

Note that while initializing all weight matrices with zero is a bad idea because of the resulting unbreakable symmetry, it is possible to set all skip connections in \eqref{eq:dense_resnet} to zero initially since the symmetry breaking will be accounted for by the weights.

\section{Related Work}

The corpus of existing literature on NAS is tremendous with new articles being released on a daily basis, see \cite{elsken2019neural} for an overview.
The majority of the established methods can be attributed the labels \emph{evolutionary}, \emph{reinforcement learning based}, or \emph{optimization based}.
Furthermore, also sparsity enforcing methods (most prominently network pruning) have strong relations to NAS.

Evolutionary methods, which construct neural networks and optimize their parameters by using, e.g., genetic algorithms, have been already proposed in the 90s \cite{angeline1994evolutionary} and are still being actively investigated \cite{elsken2018efficient,miikkulainen2019evolving}.
However, these techniques are difficult to scale to large architectures and are typically outperformed by gradient-based methods, cf., \cite{elsken2019neural}.

A very successful branch of NAS uses reinforcement learning, see, e.g., \cite{baker2016designing,zoph2016neural,zoph2018learning}.
Here, an agent designs and trains new architectures and receives rewards based on their performance or other quality measures. 
While these techniques work very well, they typically require vast amounts of computational resources since they require a full training of a possibly large architecture in every learning step.

Therefore, gradient-based optimization methods enjoy increasing popularity and oftentimes require only one training phase, which is referred to as \emph{One-Shot Neural Architecture Search}.
In \cite{liu2018darts} a differentiable architecture search method (DARTS), which continuously relaxes the search space and solves a bilevel optimization problem, is proposed.
A Bayesian optimization algorithm for determining the weights in a linear combination of network operations is studied in \cite{zhou2019bayesnas}.
In \cite{trillos2020semi,trillos2020traditional} the authors endow the set of admissible architectures with a weighted graph structure and optimize over parameters and architectures following ideas from optimal transport. 

As mentioned before NAS has strong relations to sparsity.
In particular, the lottery ticket hypothesis \cite{frankle2018lottery} states that with high probability large network architectures contain sparse sub-architectures with equal performance.
The most commonly used approach is pruning \cite{zhu2017prune,lecun1990optimal,han2015learning}, which is based on removing negligible network parameters by thresholding and can be combined with evolutionary algorithms to \emph{grow-and-prune} strategies \cite{dettmers2019sparse,liu2021,dai2019nest}.
Also sparsity-regularized training \cite{scardapane2017group} and inverse scale space approaches \cite{bungert2021bregman,huang2016split,fu2019exploring} have proven to be effective training methods for sparse neural methods, which, in addition, are amenable for mathematical analysis.
Exploiting sparsity for NAS is a more recent endeavour: \cite{wang2020you,wu2020neural} phrase the problem as bi-level optimization with sparse weighting parameters of different network operations, \cite{huang2021joint} consider a multi-objective problem, involving loss and sparsity terms.

\section{Results}\label{sec:results}

In this section we will apply the proposed NAS algorithm to different scenarios.
First, we show that it can be used to discover autoencoder-like and residual architectures for image denoising and deblurring on MNIST \cite{leCun10}.
Second, we also utilize our algorithm to unveil residual substructures of a DenseNet \cite{huang18} for the classification task on FashionMNIST.


\subsection{Denoising and Deblurring}

In this section we perform NAS for denoising and deblurring images from the MNIST dataset \cite{leCun10}.
The set consists of $60,000$ gray scale images of handwritten digits with $28\times 28$ pixels, which we split into $55,000$ images used for the training and $5,000$ test images.
For the denoising experiment we added Gaussian noise with standard deviation $0.05$.
For the deblurring experiment we first blurred the images with a Gaussian filter of size $9\times 9$ and standard deviation $5.0$, and then added the same amount of noise.
Our search space is the general network architecture \eqref{eq:dense_resnet} with $L=6$ and ReLU activations, meaning that the network has one input layer, five hidden layers, and one output layer. 
This architecture allows for feedforward networks, ResNets, Momentum ResNets, etc.
We used the algorithmic parameters as given in \cref{alg:adabreg}, with MSE loss, and a batch size of $128$.
We initially set all skip parameters to zero and set $1\%$ of all weight matrix rows non-zero, following the strategy described in \cref{sec:method}.
Using different regularization functionals $\func$ we show in the following that different architectures are discovered.

\paragraph{Autoencoders}

\begin{figure}[t]
    \centering
    {\bf Denoising} \hspace{5.5cm} {\bf Deblurring}
    
    \vspace{0.2cm}
    {\includegraphics[width=0.49\textwidth,trim=0.cm 0.cm 1.cm 1cm,clip]{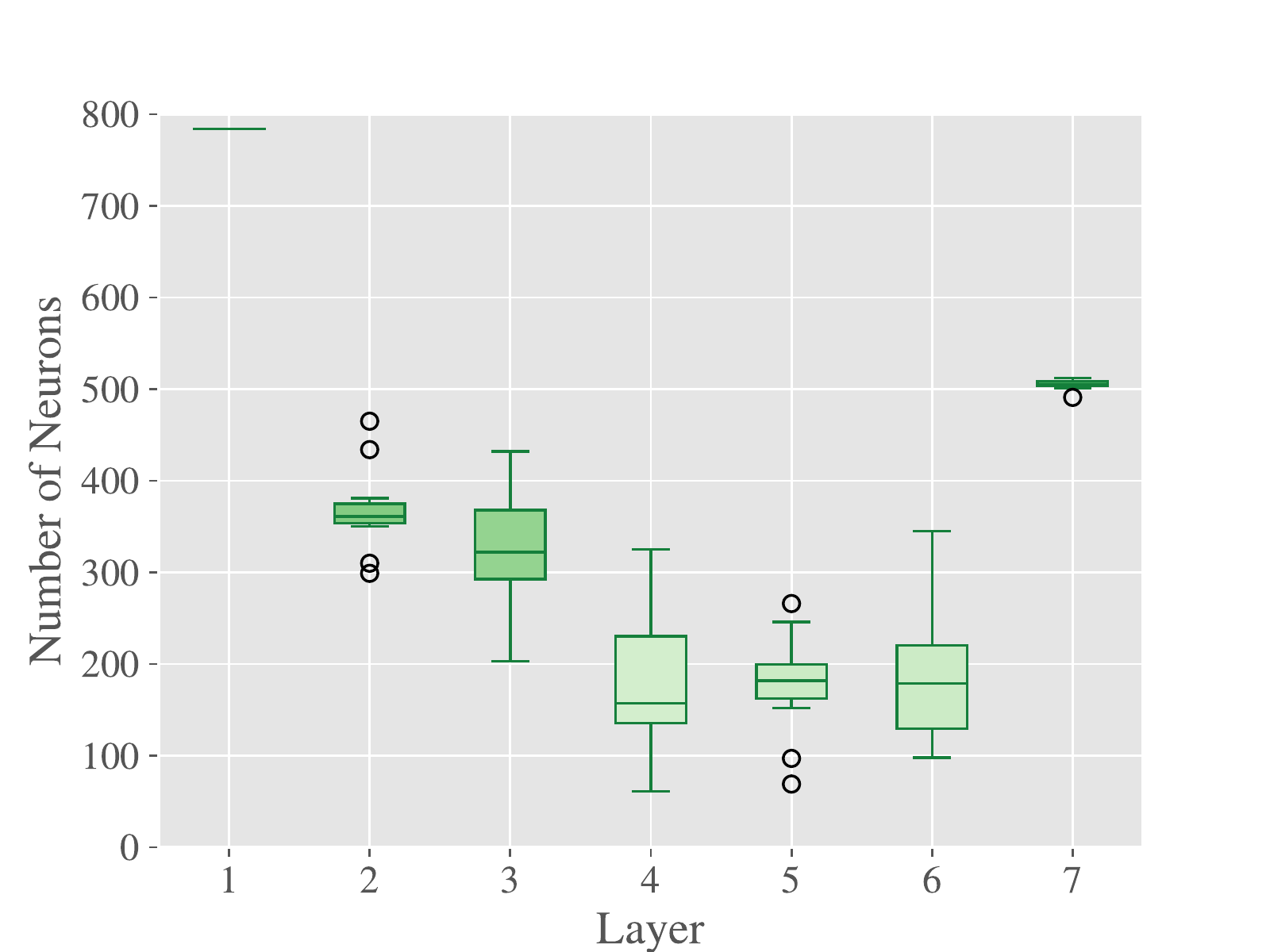}}%
    \hfill%
    \includegraphics[width=0.49\textwidth,trim=0.cm 0.cm 1.cm 1cm,clip]{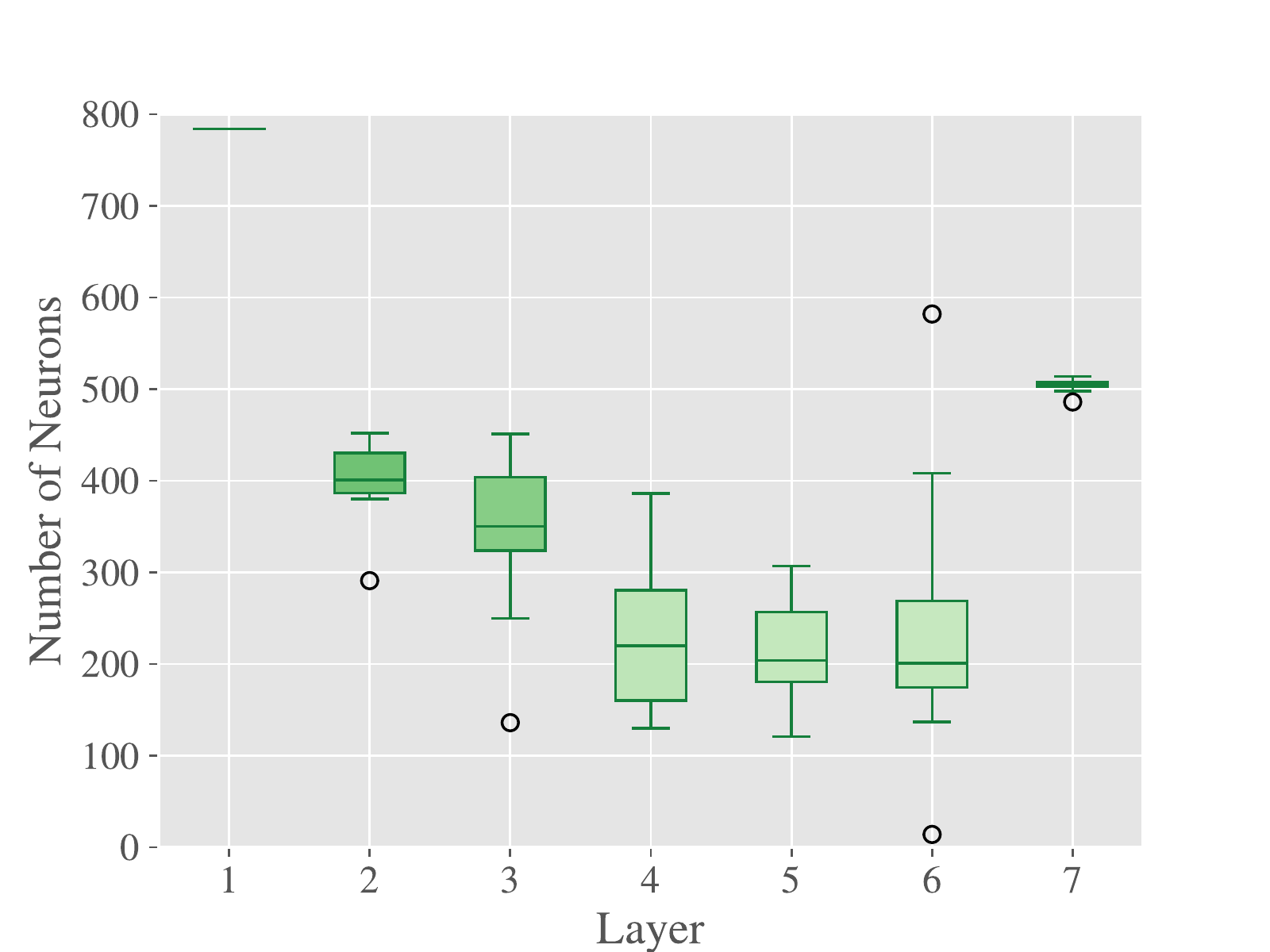}%
    \caption{Layer widths of unveiled autoencoders for denoising and deblurring.}
    \label{fig:autoencoders}
\end{figure}

In our first experiment we choose the regularizer \eqref{eq:regularizer_rows} with $\mu=0.07$ which enforces row sparsity of the weights, i.e., a small number of neurons per layer, and sets all residual connections to zero.
Hence, the search space consists of all multi-layer perceptrons with six layers of $784$ neurons each.

\cref{fig:autoencoders} shows box plots of the layer sizes of the trained networks, averaged over $10$ training runs of $100$ epochs each.
Note that we added the input layer of size $784$ manually for visualization purposes. 
Evidently, both for the denoising and the deblurring task our algorithm unveils autoencoder architectures.  
While the widths of the third to sixth layer have some spread over the runs, the average widths of the second and in particular the last layer have high statistical significance.
This suggests some degree of redundancy of the central layers, whereas the width of the output layer matches well the average amount of non-zero pixels in the MNIST data, which is around $65\%$, and hence accounts for the fact that the MNIST digits do not fill the whole image domain.

Notably, the unveiled autoencoder architectures for denoising and deblurring are very similar both qualitatively and quantitatively. 
We hypothesize that this is due to the fact that the networks' architectures stem from the challenging task of removing the noise, whereas the inversion of the Gaussian blur with a \emph{fixed} kernel can be learnt easily without significantly changing the autoencoder architecture.

\paragraph{Residual Autoencoders}

\begin{figure}[t]
    \centering
    \def\Width{0.49\textwidth}
    
    {\bf Denoising} \hspace{5.7cm} {\bf Deblurring}
    
    \vspace{0.2cm}
    {\includegraphics[width=\Width,trim=0.cm 0.cm 1.cm 1cm,clip]{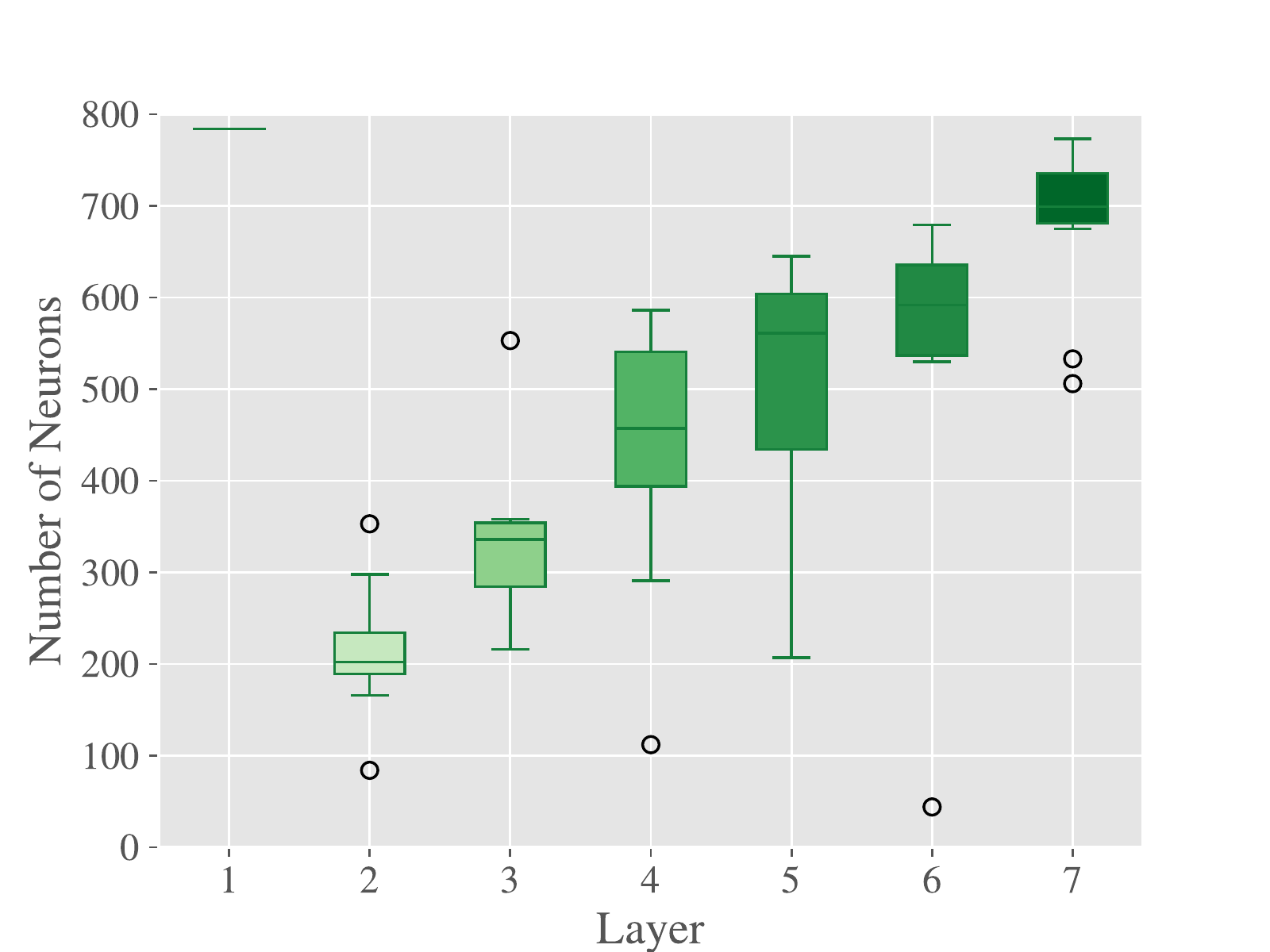}}%
    \hfill%
    {\includegraphics[width=\Width,trim=0.cm 0.cm 1.cm 1cm,clip]{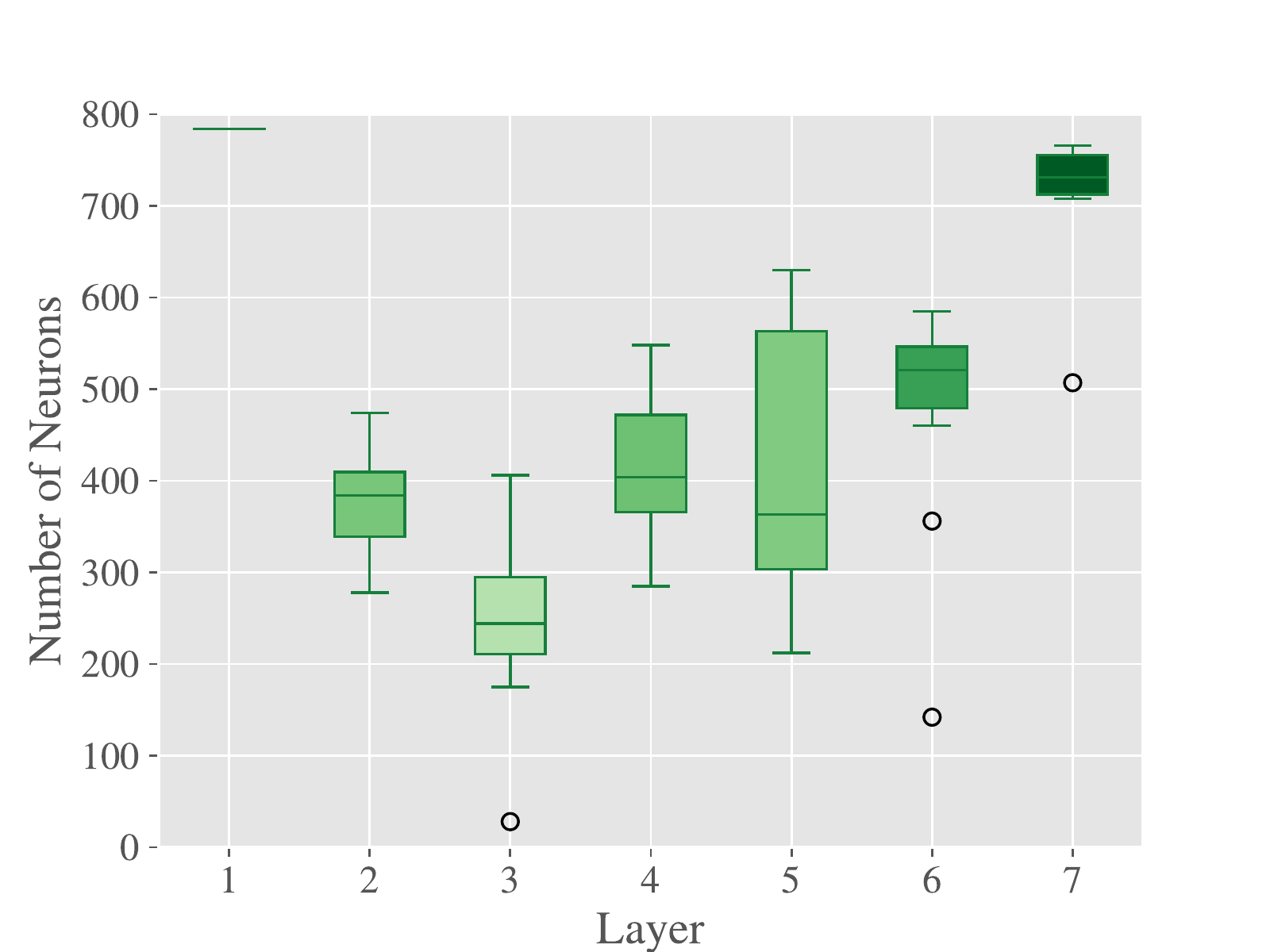}}%
    
    \includegraphics[width=\Width,trim=0.5cm .0cm 1.5cm 1.5cm,clip]{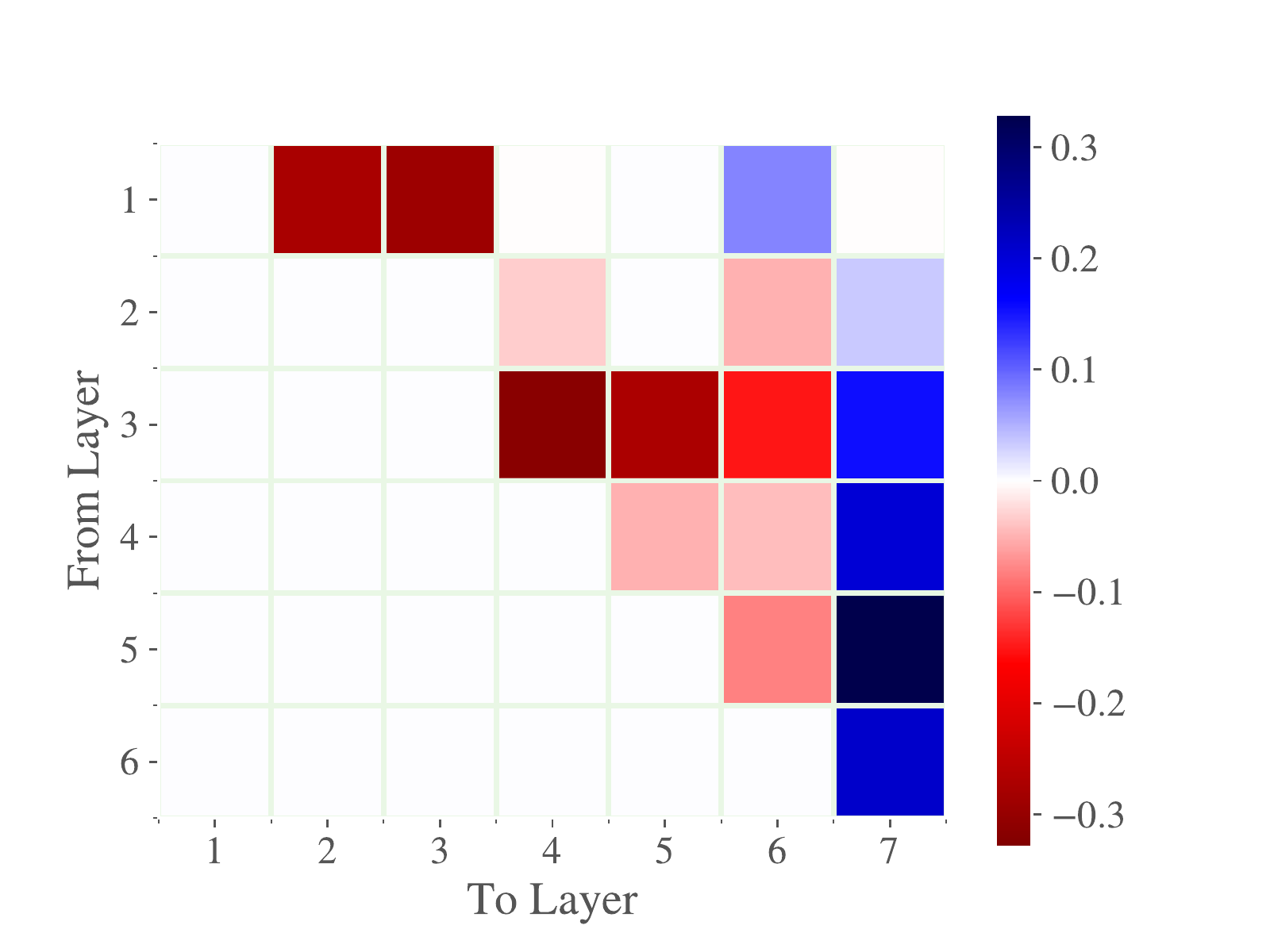}%
    \hfill%
    {\includegraphics[width=\Width,trim=0.5cm .0cm 1.5cm 1.5cm,clip]{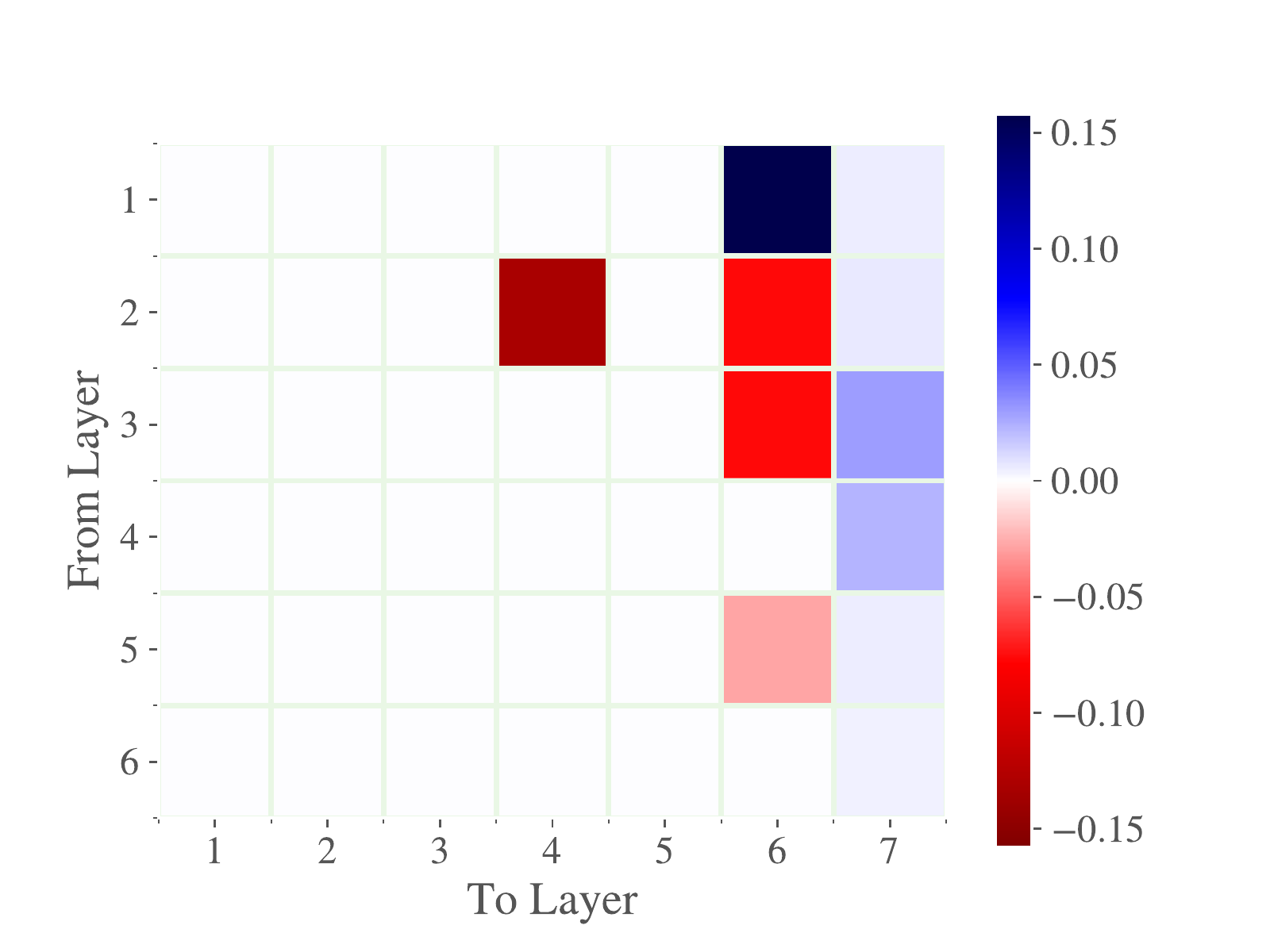}}%
    \caption{Unveiled residual autoencoder for denoising (\textbf{left}) and deblurring (\textbf{right}). \textbf{Top:} Layer widths, \textbf{Bottom:} Connectivity matrices of averaged value of skip connections, negative in red and positive in blue.}
    \label{fig:residual_autoencoders_denoising_deblurring}
\end{figure}

To enlarge the search space and thus allow for more interesting neural architectures we allow residual connections in this experiment and use the regularizing functional \eqref{eq:regularizer_skip_rows} with $\mu=0.07$, which enforces both sparse residual connections and weight rows.
The top row of \cref{fig:residual_autoencoders_denoising_deblurring} shows box plots of the layer sizes for denoising and deblurring, again averaged over $10$ training runs of $100$ epochs each.
Additionally, in the bottom row we show the average skip parameters $\tau^{(l,i)}$ in \eqref{eq:dense_resnet} as connectivity matrices, with negative values in red and positive values in blue. 
Opposed to the previous experiment without skip connections, the unveiled autoencoder structures are tilted in the sense that the coding layer of approximately $200$ neurons arises as first or second hidden layer of the network.
Also the sizes of the output layers are larger than in the previous experiment, which we attribute to the fact that the skip connections into the output layer require a larger number of neurons there to fully remove the noise or blur.

The residual connectivity matrices in the second row of \cref{fig:residual_autoencoders_denoising_deblurring} show that for both denoising and deblurring the learned architectures feature negative and positive residual parameters $\tau^{(l,i)}$. 
Interestingly, for both tasks the output layer in average receives positive residual connections from the previous layers, similar to ResNets, DenseNets, or U-Nets.
In contrast, most of the other skip parameters are negative, which is reminiscent of Momentum ResNets \cite{sander21} and typically leads to better performance than non-negative skip parameters, as observed, e.g., in \cite{benning2019deep}. 
Note that the residual networks \emph{do not} form a U-Net structure \cite{ronneberger2015u}, which would mean that most skip parameters are on the counterdiagonal of the residual connectivity matrix, connecting the first and last, second and penultimate layer, etc.
Instead the residual network for the denoising task has a triangular structure, mapping the input to the coding layer and then linearly increasing layer sizes towards the output layer, which also receives skip connections from all previous layers but the input one.

\begin{table}[t]
    \centering
    \begin{tabular}{l||c|c}
     & Denoising & Deblurring  \\
     \hhline{===}
     Autoencoders & $0.081(\pm 0.012)$ & $0.147 (\pm 0.021)$\\
     \hline
     Residual autoencoders & $\mathbf{0.005 (\pm 0.000)}$  & $\mathbf{0.112 (\pm 0.006)}$
\end{tabular}
    \caption{Average MSE loss ($\pm$ standard deviation) on MNIST test data set.}
    \label{tab:loss_MNIST}
\end{table}

Finally, we compare the different architectures in terms of their MSE loss on the MNIST testing data set. 
\cref{tab:loss_MNIST} shows the average MSE losses and standard deviations of the unveiled autoencoders and residual autoencoders for both the denoising and the deblurring tasks.
Here, the residual architectures have a significantly lower loss than the autoencoder like structures. 
The improvement is especially high for the denoising task where the loss drops more than one order of magnitude when allowing for residual connections.

\subsection{Classification}

In this experiment we perform classification on the Fashion-MNIST dataset, which is a more challenging drop-in replacement of MNIST.
We use the same train-test splits and algorithmic parameters as before.
The loss function is chosen as cross-entropy loss.

The search space is constituted by a DenseNet architecture \cite{huang18}, which consists of a 5-layer dense block with a growth rate of $k=12$, followed by a convolutional and a linear layer.
As regularizer we choose a group norm on all convolutional kernels,
\begin{align*}
    \func(\param) = \mu\sum_{\iota\in I_\mathrm{conv}} \sqrt{s}\norm{K^{(\iota)}}_F,
\end{align*}
where $I_\mathrm{conv}$ is the index set of the kernels, $s$ denotes the kernel size, and $\norm{\cdot}_F$ is the Frobenius norm. 
The regularization parameter was chosen as $\mu=0.05$.
In this case the skip connections are realized as concatenations, 
where the $l$-th layer of a dense block with growth rate $k$ receives an input 
$x^{(l)}\in\R^{l\cdot k, n,m}$ and outputs a vector $x^{(l+1)}$, which is given by
\begin{alignat*}{2}
x^{(l+1)}_{j,\bullet} &= x^{(l)}_{j,\bullet},\qquad &&j\in\{0,\ldots,l\cdot k - 1\},\\
x^{(l+1)}_{j + l\cdot k,\bullet} &= \sum_{i=0}^{l\cdot k} K^{(l,i,j)} \ast x^{(l)}_{i,\bullet},\qquad &&j\in\{0,\ldots,k-1\}.
\end{alignat*}
In this formulation, the 
kernel matrices for $j\leq (l-1)\cdot k$ are a generalization of the 
skip connections in \eqref{eq:dense_resnet} since they multiply the outputs of previous layers. In particular one can compute the values
\begin{align*}
\tau^{(l,r)} := \sum_{i=r\cdot k}^{(r+1)\cdot k} \sum_{j=1}^{k} \norm{K^{(l,i,j)}}
\end{align*}
for $r\leq l$, which act as strength of the residual connections from layer $r$ to layer $l$.

\begin{figure}[thb]
    \centering
    \def\Width{0.45\textwidth}
    {\includegraphics[width=\Width,trim=0.5cm 0cm 1cm 1.cm,clip]{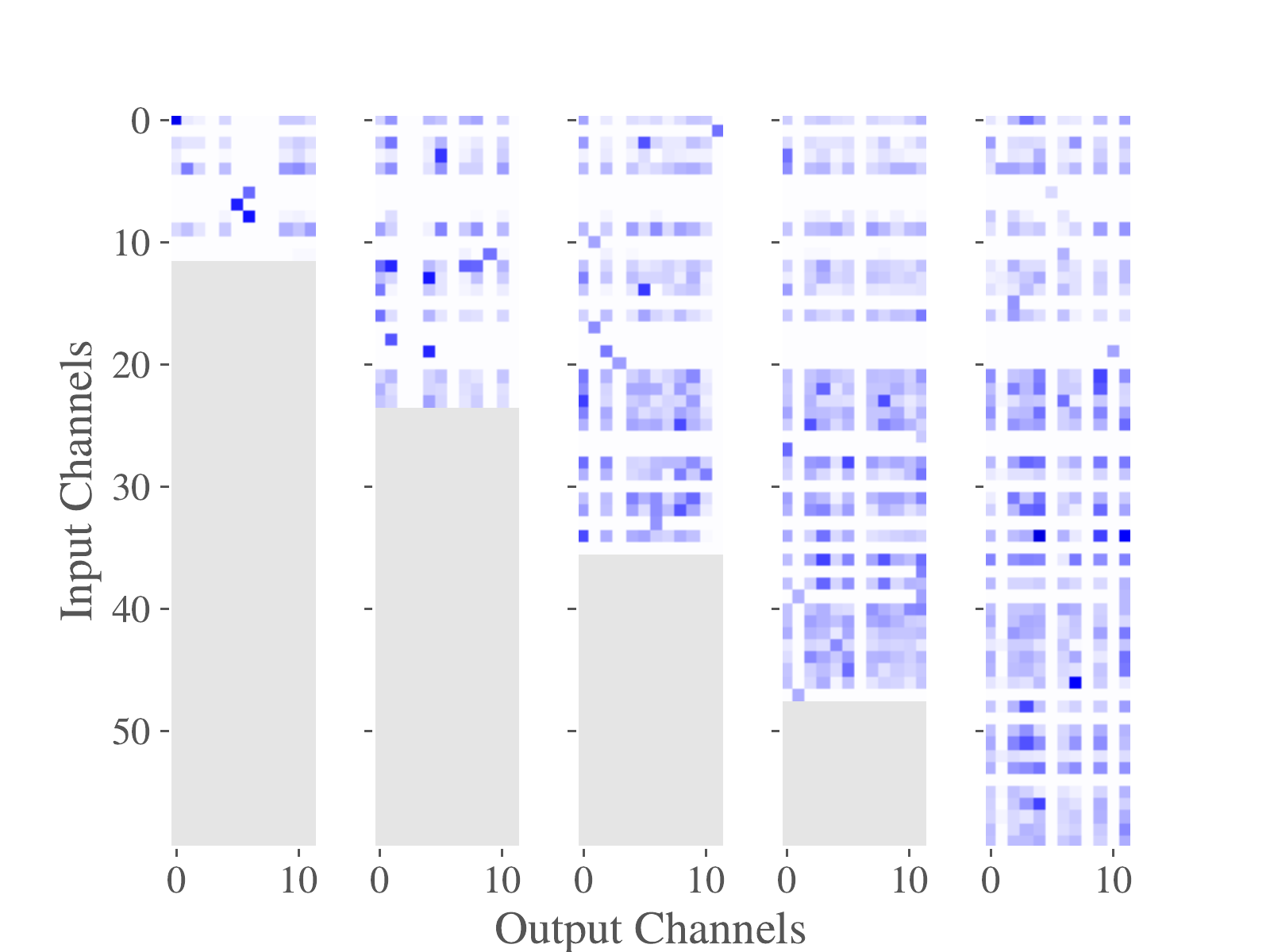}}%
    \hfill%
    {\includegraphics[width=\Width,trim=0.7cm 0.5cm 2cm 1.cm,clip]{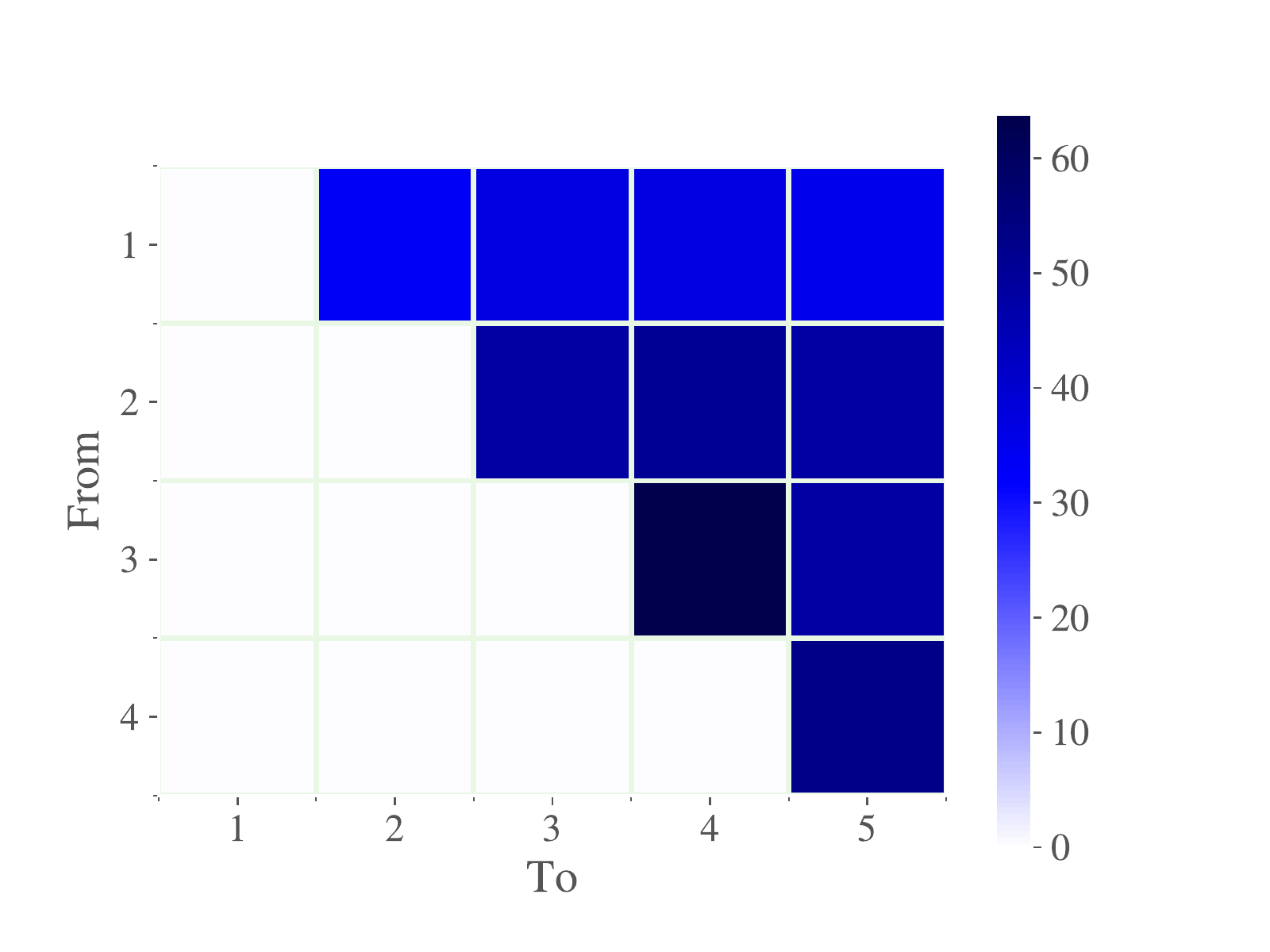}}%
    \caption{\textbf{Left:} Residual kernel matrix for individual channels. \textbf{Right:} Residual kernel matrix summed over all channels in a layer.}
    \label{fig:densenet}
\end{figure}

The left part of \cref{fig:densenet} shows the norms of all kernels and the right part demonstrates the residual connections $\tau^{(l,r)}$, as defined above.
Similar to the previous experiments, which however dealt with the completely different tasks of denoising and deblurring, the last layer receives most skip connections.
This again indicates a ``triangular'' structure of the unveiled network.
Furthermore, from the left image in \cref{fig:densenet} we note that some of the input and output channels are not used by the network.

\section{Discussion and Future Work}\label{sec:discussion}

In this paper we proposed a new one-shot approach for optimization-based Neural Architecture Search (NAS) based on an inverse scale space method.
We initialize a sparse neural network within a defined search space and the proposed algorithm performs Bregman iterations to add only relevant parameters to the sparse network.
This strategies allows the neural network to evolve to a sparse neural architecture that is well-designed for a given task.
Our experiments with different application tasks demonstrated that choosing different regularization functionals yield interesting neural architectures, which are also known to be successful from the literature, e.g., autoencoders and residual networks.

Some limitations of our method are the following:
First, it requires a network superstructure which can become infeasibly large when supposed to cover, e.g., convolutional layers with hundreds of channels.
Second, it contains two hyperparameters $\delta>0$ and $\mu>0$, which might require tuning. 
For Bregman iterations of convex losses $\delta$ can be shown to have no influence \cite{yin2010analysis}, which is why chose $\delta=1$ in this work.
The second parameter $\mu>0$ steers the amount of thresholding applied to the subgradients in \cref{alg:adabreg} and has greater influence on the sparsity of the resulting networks~\cite{bungert2021bregman}.

The proposed method has great potential for unveiling unexplored architectures for real world applications.
Hence, in future work we aim to perform computations on larger neural architectures, possibly combining it with reinforcement learning-based NAS.
Furthermore, we will investigate if the proposed method can be translated to meta neural architecture search \cite{wang2020m}, in which suitable architecture blocks (known as motifs) are automatically combined by the inverse scale space approach.
This would mean that neural architecture design becomes significantly less time-consuming than manually testing combinations of different building blocks of neural networks, e.g., pooling, dropout, dense, or residual layers.

\printbibliography

\end{document}